\documentclass[11pt]{article}
\usepackage{times} 
\usepackage[top=2.5cm, bottom=2.5cm, left=2.5cm, right=2.5cm]{geometry}
\usepackage[square]{natbib}
\usepackage{amsmath}
\usepackage{amssymb}
\usepackage{amsfonts}
\usepackage{epsfig}
\usepackage{url}

\usepackage{microtype}

\usepackage{latexsym}

\title{Visualizing Topics with Multi-Word Expressions}

\author{David M. Blei \\
  Department of Computer Science \\
  Princeton University \\
  Princeton, NJ 08540, USA\\
  {\tt blei@cs.princeton.edu} \and
  John D. Lafferty \\
  Computer Science Department \\
  Machine Learning Department\\
  Carnegie Mellon University \\
  Pittsburgh, PA 15213, USA\\
  {\tt lafferty@cs.cmu.edu}}

\usepackage{amsmath}
\usepackage{amssymb}
\usepackage{amsfonts}
\usepackage{epsfig}

\usepackage[varg]{txfonts}

\newcommand{\dir}{\textrm{Dirichlet}}
\newcommand{\mult}{\textrm{Multinomial}}
\newcommand{\myeq}[1]{equation~\eqref{eq:#1}}
\newcommand{\mysec}[1]{Section~\ref{sec:#1}}
\newcommand{\myfig}[1]{Figure~\ref{fig:#1}}
\newcommand{\hist}{h}
\newcommand{\new}{\textrm{new}}
\newcommand{\g}{\,|\,}

\newcommand{\lr}{\textrm{LR}}
\newcommand{\pinew}{\pi^{\textrm{new}}}
\newcommand{\ie}{\textit{i.e.}}
\long\def\ignore#1{}

\newenvironment{packed_enumerate}{
  \begin{enumerate}
    \setlength{\topsep}{0pt}
    \setlength{\itemsep}{6pt}
    \setlength{\parskip}{0pt}
    \setlength{\parsep}{0pt}
}{\end{enumerate}}

\begin{document}

\maketitle

\begin{abstract}
  We describe a new method for visualizing topics, the distributions
  over terms that are automatically extracted from large text corpora
  using latent variable models.  Our method finds significant
  $n$-grams related to a topic, which are then used to help understand
  and interpret the underlying distribution.  Compared with the usual
  visualization, which simply lists the most probable topical terms,
  the multi-word expressions provide a better intuitive impression for
  what a topic is ``about.''  Our approach is based on a language
  model of arbitrary length expressions, for which we develop a new
  methodology based on nested permutation tests to find significant
  phrases.  We show that this method outperforms the more standard use
  of $\chi^2$ and likelihood ratio tests.  We illustrate the topic
  presentations on corpora of scientific abstracts and news articles.
\end{abstract}

\section{Introduction}
\thispagestyle{empty}

Topic models are hierarchical Bayesian models of document collections
that explain an observed corpus with a small set of distributions over
terms.  When fit to a corpus, these distributions tend to correspond
to intuitive notions of the topics or themes that pervade the
documents.  Topic models have emerged as a powerful tool for
unsupervised analysis of text.  They have been extended for
authorship~\cite{Rosen-Zvi:2004}, citation~\cite{McCallum:2007}, and
discourse segmentation~\cite{Purver:2006}.  Review articles of topic
modeling provide further applications~\cite{Griffiths:2006}.

The idea behind topic modeling is to imagine a probabilistic process
by which both a hidden thematic structure and observed collection of
documents arises.  Given the observed collection, one then
``reverses'' this process to determine the posterior distribution of
the hidden thematic structure.  Topic models build on and were
inspired by techniques like latent semantic analysis
(LSA)~\cite{Deerwester:1990} and probabilistic latent semantic
analysis (pLSA)~\cite{Hofmann:1999b}.  However, LSA and pLSA do not
embody generative probabilistic processes.  By adopting a fully
generative model, topic models such as latent Dirichlet allocation
exhibit better generalization and are easily
extendable~\citep{Blei:2003b}.

Once they are fit to a corpus, it is of interest to visualize the
topics.  These visualizations provide landmark descriptive statistics
for understanding, exploring, and navigating through an otherwise
unorganized collection of documents~\cite{Mimno:2007}.  Typically, one
visualizes each topic by simply listing the terms in order of
decreasing probability.  While a person can usually peruse these lists
and intuit ``meanings'' of the topics, such visualizations can be
unsatisfying.  Single terms are often part of indicative phrases,
which are lost in a simple unigram representation.  An alternative is
to fit a more complicated
model~\citep{Girolami:2004,Wallach:2006,Wang:2007}, but then one loses
the computational advantage and statistical simplicity of unigram
topic modeling.

In this paper we introduce a new method for visualizing unigram topic
models.  In our approach, the model is first fit as usual, and then
the posterior distribution is used to annotate each word occurrence of
the corpus with its most probable topic.  With this annotated corpus, we
carry out a statistical co-occurrence analysis to extract the most
significant $n$-grams for each topic.  The resulting multi-term
phrases are combined with the unigram lists to give a visualization
that offers a better intuitive impression for what a topic is about.
We call the resulting visualizations \textit{turbo topics}, as
suggested by the manner in which the method recycles the output of
estimation to build a more powerful presentation of the model.

As part of this procedure, we developed a new algorithm for finding
multi-word expressions.  Our method uses a back-off language model
defined for arbitrary length expressions~\cite{Katz:1987}, and
recursively employs the distribution-free permutation test to find
significant phrases.  In contrast, previous methods of finding
multi-word expressions rely on a test statistic derived from a
multinomial contingency table and, in most cases, appeal to the
asymptotic distribution of that statistic~\cite{Manning:1999}.  We
show that the permutation test works better in small sample settings,
such as when we restrict our attention to topical terms, and the
back-off model allows for finding multi-word phrases within a
well-defined language model.

We describe turbo topics in \mysec{turbotopics} and our new method of
finding multi-word expressions in \mysec{model}.  In \mysec{results},
we evaluate on simulated data and illustrate improved topic
visualization with two real-world corpora.





\section{Turbo Topics}
\label{sec:turbotopics}

\begin{figure*}[t]
\begin{center}
  \includegraphics[width=\textwidth]{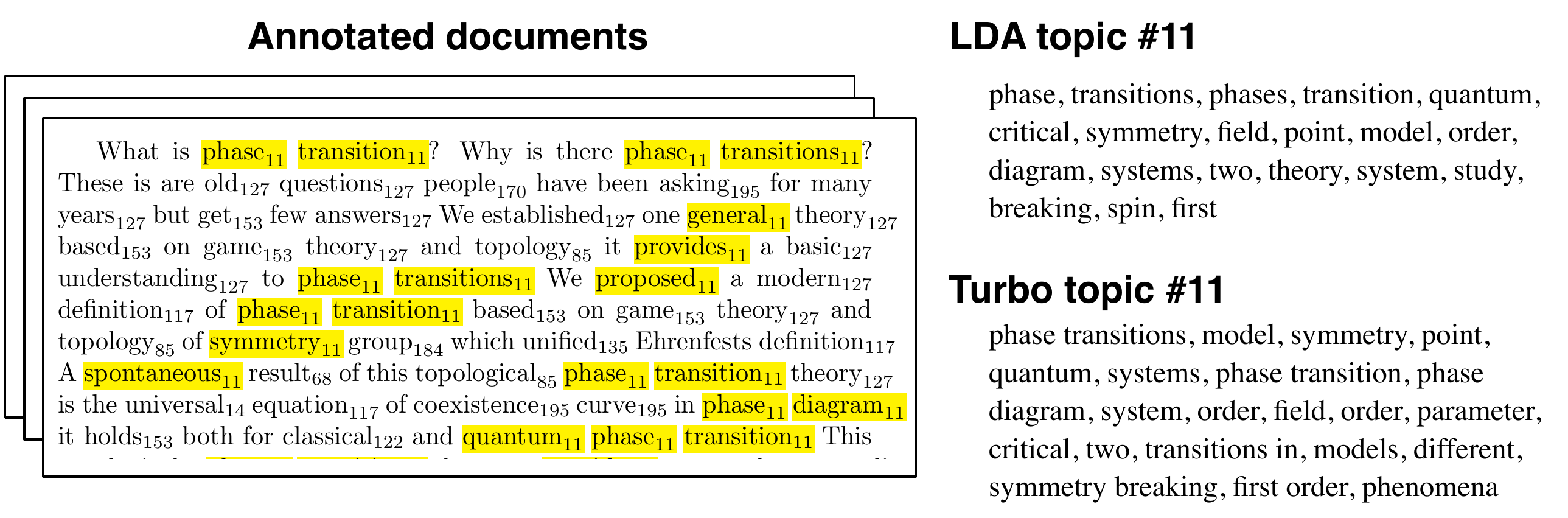}
\end{center}
\caption{\label{fig:method} An illustration of the turbo topics
  strategy.  We first estimate an LDA topic model (under the word
  exchangeability assumption).  We next annotate each word in the
  original corpus with its most likely posterior topic.  This is
  illustrated at left in the subscript on each word and with topic 11
  highlighted in yellow.  We run a hypothesis testing procedure over
  the annotated corpus to identify significant words that appear to
  the left or right of a word or phrase labeled with a given topic.
  This procedure is carried out recursively, until no more significant
  phrases are found.  At right we illustrate the original top words
  from topic 11, and those find by the turbo topics strategy.  Phrases
  like ``phase diagram,'' ``symmetry breaking,'' and ``first order''
  are found by the procedure.  More topics are illustrated in
  \myfig{huff}.}
\end{figure*}

We first review latent Dirichlet allocation (LDA), a commonly used
building block for topic models~\citep{Blei:2003b}.  We then describe
our algorithm for finding multi-word expressions to visualize topics.

\paragraph{Latent Dirichlet allocation.}  LDA models documents as
arising from multiple topics, where a \textit{topic} is defined to be
a distribution over a fixed vocabulary of terms~\citep{Blei:2003b}.
Specifically, LDA assumes that $K$ topics are associated with a
collection, and that each document exhibits these topics with
different proportions.  This is often a natural assumption because
documents tend to be heterogeneous.  Each might combine a subset of
themes that permeate the collection as a whole.

LDA is a hidden variable model where the observed data are the words
of each document and the hidden variables are the latent topical
structure, \ie, the topics themselves and how each document exhibits
them.  Given a collection, the \textit{posterior distribution} of the
hidden variables given the words of the documents provides a
probabilistic decomposition of the documents into topics.

The statistical assumptions underlying LDA can be understood by its
probabilistic generative process, the random process that is assumed
to have produced the observed data.  Let $K$ be a specified number of
topics, $V$ the size of the vocabulary, $\alpha$ a positive
$K$-vector, and $\eta$ a scalar.  LDA assumes that the collection
arises as follows.

For each of the $K$ topics draw a distribution over words,
\[ \beta_{k} \sim \dir_V(\eta), \] where $\dir_V$ denotes the
Dirichlet distribution on the $V-1$ dimensional probability simplex.
For each document $d$ draw a vector of topic proportions
\[\theta_{d} \sim \dir_K(\alpha).\]
Finally, to select the $i$th word in the document, first draw a topic
assignment from the topic proportions,
\[Z_{d,i} \g \theta_d \sim \mult(\theta_{d})\] and then draw the word
from the chosen topic
\[W_{d,i} \g z_{d,i} \sim \mult(\beta_{z_{d,i}}).\] This process
specifies how the latent variables interact to produce the observed
collection.

Finding the posterior distribution of the hidden variables is akin to
``reversing'' this process given a corpus $\{\vec{w}_d\}_{d=1}^{D}$.
The posterior $p(\beta_{1:K}, \theta_{1:D}, \vec{z}_{1:D} \g
\vec{w}_{1:D}, \alpha, \eta)$ provides a probabilistic decomposition
of the corpus into its topics $\beta_{1:K}$.  Each document exhibits
multiple topics via its topic proportions $\theta_d$; the words of
each document are assigned to specific topics $z_{d,i}$.

However, this posterior is intractable to compute; the central
computational problem in topic modeling is to approximate it.
Efficient approximation algorithms include Markov chain Monte Carlo
sampling~\citep{Griffiths:2006} and variational
methods~\citep{Minka:2002,Blei:2003b,Teh:2006}.  Here we use
mean-field variational inference~\citep{Blei:2003b}, though our
methods can be used with any algorithm for approximate inference.

\paragraph{Multi-word expressions for topic visualization.}
\label{sec:strategy}





The topics in LDA are unigram distributions over words, and the LDA
model is exchangeable at the word level.  This means that if the words
in the corpus were shuffled within their documents then exactly the
same inferences would result.  Arguably, this is a reasonable
assumption for topic modeling.  When presented with a jumbled
document, a human can often discern the thematic content of the text,
even if he or she cannot reconstruct the detailed flow of the
presentation.  Exchangeable word models like LDA are simpler and offer
computational advantages over more complex models that take word order
into account~\citep{Girolami:2004,Wallach:2006,Wang:2007}.

Our goal is to enhance the \textit{interpretation} of the model rather
than the model itself, preserving the advantages of exchangeable
modeling while attaining the expressive visualizations of $n$-gram
modeling.  Thus, our focus is on analyzing the posterior distribution
of the topic structure of a corpus to determine phrases indicative of
each of the topics.  To do so, we first use the posterior of the topic
variables $Z_{d,i}$ to assign topics to words.  Then, based on the
original order of the words in the documents, we use the annotated
words to find the significant topical $n$-grams that stem from them.

Our strategy is as follows.
\begin{packed_enumerate}
\item Estimate an LDA topic model with $K$ topics.  This results in a
  posterior for topics, topic proportions, and per-word topic
  assignments.

\item Using the posterior, annotate each word in the original corpus
  with a topic assignment.  This yields an ordered sequence of
  word-topic pairs
  \[(w_1,z_1), (w_2,z_2), (w_3,z_3), (w_4,z_4) , \ldots\]

\item Given a word or phrase $w$ and topic $z$ of interest, run a
  hypothesis testing procedure to identify words $v$ that are likely
  to precede or follow $w$ when it is labeled as belonging to topic
  $z$.

\item Repeat step 3 until no significant phrases are added.
\end{packed_enumerate}
This is illustrated in \myfig{method}.

In step 2, the sequence $w_1, w_2, w_3,\ldots $ denotes the original
text that comprises the corpus.  This step annotates each word $w_i$
in the text with a topic assignment $z_i$, unless the word was removed
by pre-processing (e.g., a stop word).  The topic assigned to the word
is determined by the posterior, and is document specific.  Thus, the
word ``fly'' in one document might be annotated by a topic about
insects; the same word in another topic might be annotated by a topic
about airplanes.  Topic models capture polysemy, in the sense that
they can assign the same term to different topics in different
document contexts~\cite{Griffiths:2006}.  Once the words are annotated
with topic assignments, document boundaries are ignored.

Step 3 results in bigrams $(w,v)$ or $(v,w)$ for a given topic.  Note
that $v$ may or may not have been assigned topic $z$. For instance,
consider a topic focused on movies, and consider the movie title ``Sex
in the City.''  The terms ``the'' and ``and'' are stop words, which
are not assigned to any topic; the term ``city'' may not be very
relevant to the movies topic.  However, if that topic assigns high
probability to the term ``sex,'' then our method will be able to
identify the movie title because of the repeated context in which it
appears.  (See Figure~\ref{fig:huff}.)

\section{Recursive Permutation Tests For Multi-Word Expressions}
\label{sec:model}

A key component in turbo topics is the method for recursively
identifying significant $n$-grams in a sequence of words.  We describe
our novel solution to this problem, appropriate for the sparse
settings that arise in topic models.

\paragraph{A language model.}  Consider a corpus of $N$ words. Under
an arbitrary language model, the log likelihood is
\begin{equation}
\label{eq:likelihood}
  \log p(w) =
  \textstyle \sum_{n=1}^{N} \log p(w_n \g w_1, \ldots, w_{n-1}).
\end{equation}

In parameterizing this model, we can consider two extremes.  On one
extreme, the fully parameterized model contains a conditional
distribution of words given each possible history.  However, this
model is computationally intractable because it requires specifying
$O(V^N)$ parameters.  Moreover, it is statistically unreliable; there
will typically not be enough data to support maximum likelihood
estimates.  At the other extreme, the unigram model posits that each
word is independent of its history, $p(w_n \g w_1, \ldots, w_{n-1}) =
p(w_n)$.  This model is efficient to estimate, but cannot capture
dependencies between words.


We adopt a middle ground between these models that sparsely
parameterizes the full model using word histories of varying lengths.
For example, consider a model where all words follow a unigram
distribution except for those words that follow the word ``new.''
Among words following ``new,'' some words, such as ``house,''
essentially follow their unigram distribution, while others, such as
``york'' or ``jersey,'' are endowed with ``new''-specific
probabilities.

Our challenge is to determine which $n$-grams, such as ``new york,''
should be given special probabilities $p(\textrm{``new''})
p(\textrm{``york''} \g \textrm{``new''})$ and which should be modeled
by products of their unigram probabilities.  Then, to visualize the
distribution, we order by probability the collection of $n$-grams
represented in the model.

Evaluating the likelihood in \myeq{likelihood} requires a distribution
over words conditioned on an arbitrary history of previous words.
Denote a length $n$ history by $w_{1:n}$ and let $S_{w_{1:n}}$ be a
set of words that are governed by history-specific probabilities.  To
continue the example, if the history is ``new'' then
$S_{\textrm{``new"}}$ might be \{``york'', ``jersey'',
``hampshire''\}.  The conditional distribution over words is
\begin{equation}
  \label{eq:model}
  p(w_{n+1} \g w_{1:n}) =
\begin{cases}
  \pi_{w_{n+1} \g w_{1:n}} & \textrm{if $w_{n+1} \in S_{w_{1:n}}$} \\
  \gamma_{w_{1:n}} p(w_{n+1} \g w_{2:n}) & \textrm{otherwise.}
\end{cases}
\end{equation}
The constant $\gamma_{w_{1:n}}$ ensures that the distribution sums to
one,
\begin{equation}
  \label{eq:scaling}
  \gamma_{w_{1:n}} = \frac{1 - \textstyle \sum_{v \in S_{w_{1:n}}} \pi_{v \g u}}
  {1 - \sum_{v \in S_{w_{1:n}}} p(v \g w_{2:n})}.
\end{equation}
This is a back-off model with a sparsely represented set of
conditional distributions.  Note that if $S_{w_{1:n}} = \{\}$ is
empty, then $w_{1:n}$ is not endowed with a specific conditional
distribution.  In this case, $\gamma_{w_{1:n}} = 1$ and \myeq{model}
gives that $p(w \g w_{1:n}) = p(w \g w_{2:n})$.  This type of model
has been investigated in the speech recognition and language modeling
literature~\cite{Katz:1987,Seymore:1996,Stolcke:1998}.


\paragraph{Expanding the model with likelihood ratios.}  We now
describe how we search through the space of sparsity patterns for the
parameters of the language model.  Finding a good set of parameters
amounts to identifying the set $S_\hist$ for each word history $\hist$
in a corpus.  Beginning with a unigram model, where $S_{\hist} = \{\}$
for all histories, our approach is to greedily determine the words
best governed by bigram probabilities.  We then apply this procedure
recursively to find higher order dependencies.

Consider a sparse bigram model, a model with word histories of at most
one word, and consider a single pair of words $u$ and $v$ such that $v
\notin S_u$.  To determine whether to add a bigram parameter $\pi_{v
  \g u}$ to the model, we compute the log likelihood ratio of the
expanded model to the unexpanded model, \ie, the log of the
probability of the data under the expanded model divided by the
probability of the data under the unexpanded model.  The expanded
model is the model with $S_u \leftarrow S_u \cup \{v\}$ and
conditional probabilities following \myeq{model}. It contains one new
parameter $\pinew_{v \g u}$, and a different estimate of the parameter
$\pinew_v$.  All other parameters are identical in both models.

For the likelihood ratio, the only probabilities that change between
the two models are for the words that follow $u$ and for all instances
of $v$.  The probability of instances of $v$ following $u$ change from
their unigram probability $\pi_v$ to their bigram probability
$\pinew_{v \g u}$; the probability of other instances of $v$ changes
from $\pi_v$ to $\pinew_v$. Words following $u$ that are not in $S_u$
are governed by their unigram probabilities; however, they are scaled
differently in the expanded model. The scaling constant of
\myeq{scaling} uses both $\pinew_{v \g u}$ and $\pinew_{v}$.

Thus, the log likelihood ratio of the expanded model to the unexpanded
model is
\begin{align}
  \label{eq:score}
  \nonumber \lr_{uv} = & n_{uv} \log \pinew_{v \g u} + (n_v - n_{uv})
  \log \pi^{\textrm{new}}_{v} \\
  & + \textstyle \sum_{v' \in S^c_u \slash v} n_{uv'} \log
  (\pi_{v'}
  \gamma^{\new}_u) \\
  & \nonumber - n_{uv} \log \pi_v - \textstyle \sum_{v' \in S^c_u
    \slash v} n_{uv'} \log (\pi_{v'} \gamma_u).
\end{align}
The first two lines are the log likelihood under the expanded model;
the third line is the negative log likelihood under the unexpanded
model.  All parameters are computed as normalized
counts~\citep{Katz:1987}.  A value of $\lr_{uv}$ above zero indicates
that the expanded model fits the data better than the unigram model.
This quantity is closely related to an entropy-based
score~\cite{Stolcke:1998}.


For simplicity, we have described the likelihood ratio for expanding a
unigram model to a sparse bigram model.  This methodology can be
generalized to word histories of arbitrary length, and the resulting
likelihood ratios can be used to assess expansions beyond bigrams.  In
assessing such expansions, we compute the ratio of the log likelihoods
for a model with history $\hist$ and a model with expanded history
$\hist \cup \{v\}$.  The unexpanded model's back-off probabilities are
for words given the original history.

\paragraph{Recursive permutation tests.}

With the likelihood ratio in hand, our next task is to develop an
algorithm for building up a model from the unigram parameterization.
Our algorithm adds parameters as needed by carrying out a sequence of
hypothesis tests.

Given a previous word $u$, consider the best candidate word that is
not yet modeled as a bigram,
\begin{equation*}
  v^* = \arg \max_{v \notin S_u} \lr_{uv}.
\end{equation*}
Suppose that $\lr_{uv^*}$ is greater than zero.  When is this
significant, and when it is an artifact of a small data sample?

We answer this question with a permutation
test~\citep{Pitman:1937,Good:2000}.  The permutation test determines
the significance of a score, as in \myeq{score}, that measures the
degree of dependence between two random variables.  To decide whether
a log likelihood ratio is positive, we shuffle the data in such a way
as to remove the added dependence, but retaining the previously
modeled dependencies.  We then compute the same log likelihood ratio
on the shuffled data.  A score computed from shuffled data that is
greater than the score we are testing is evidence that the true score
is not significant, because it arose in a data set where any such
dependency has been removed by design.  Repeating this process, the
proportion of scores under shuffled data greater than the quantity in
question provides a p-value for its significance.

Many hypothesis tests rely on assumptions about the asymptotic
distribution of the test statistic.  The permutation test relies on no
such asymptotic assumption, and is thus particularly suited to sparse
data settings.  In natural language processing, permutation tests have
been used for word collocations in multinomial
models~\citep{Pedersen:1996} and for bilingual
associations~\citep{Moore:2004}.  They have not been developed for the
back-off language models that we consider.

Returning to our task, recall that a positive likelihood ratio
$\lr_{uv^*}$ indicates that $u$ and $v^*$ are dependent in the joint
distribution.  To perform the test, we repeatedly shuffle all the
words (within and across documents), but retain the sequences of words
that are currently modeled.  This removes dependencies between $u$ and
terms that are not in $S_u$, while retaining the other dependencies
that are already assumed.  In the permuted data, if there is a $v$
such that $\lr^{\textrm{permuted}}_{uv}$ is greater than $\lr_{uv^*}$,
then this is an indication that our likelihood ratio score is not
significant.


We perform the following procedure to find the set of words $S_u$ with
which to endow special $u$-specific probabilities,
\begin{packed_enumerate}
\item Set $v^* = \arg \max_{v \notin S_u} \lr_{uv}$.
\item Sample $M$ permutations of the data that respect the current
  estimate of $S_u$.  Compute
  \begin{equation*}
    \textrm{p-value}_{uv^*} \approx
    \frac{1}{M} \left(\textrm{\# scores} > \lr_{uv^{*}}\right)
  \end{equation*}
\item If $\textrm{p-value}_{uv^{*}}$ is less than the desired
  threshold then add $u$ to $S_u$ and repeat.
\end{packed_enumerate}

\paragraph{Intuitions and previous approaches.}

There are two primary differences between this approach and previous
approaches for finding phrases.  First, by testing the
\textit{maximum} log likelihood ratio, we address the issue of finding
multiple related collocations rooted in a single word.  If we
simultaneously tested each possible expansion, then the hypothesis
tests would be dependent on each other.  Traditional phrase finding
algorithms are based on a multinomial contingency table and subsequent
hypothesis test~\citep{Manning:1999}.  They test all collocations
simultaneously, without accounting for the bias that this introduces.

This point can be made more concrete with our running example.  Suppose
``new'' occurs 10,000 times in the corpus, the word ``york'' follows
it 6000 times, and the word ``jersey'' follows it 3,000 times.  Once we
take ``york'' into account, ``new jersey'' can occur at most 4000
times.  By not accounting for the the instances of ``new york'' we see
that 3000 out of 4000 is a strong signal of a bigram.  Leaving in
``new york,'' as we would if we simultaneously tested both words, we
would find that 3000 out of 10,000 is not as strong a signal.

Second, and more importantly, our method is based on expanding the
sparsity pattern of the parameters of the model in \myeq{model}.  When
a significant word collocation is found, we expand the model to share
fewer parameters; but the resulting expanded model is still a valid
model.  We can thus apply the hypothesis test recursively.



In traditional algorithms for finding collocations, there is no
principled way to expand a model once a significant pair of words is
found.  For example, suppose a traditional algorithm finds ``new
york'' to be a significant bigram.  One can try to add ``new york'' to
the vocabulary of the multinomial model.  However, the resulting
distribution, with probabilities for ``new'', ``york'' and ``new
york'', will then have two ways of generating the bigram ``new york.''
(One way is to generate the two words independently; the other is to
generate the added bigram.)  It is not clear how to account for
observed text with such a distribution.


\section{Empirical Results}

\begin{figure*}[t]
   \includegraphics[width=\textwidth]{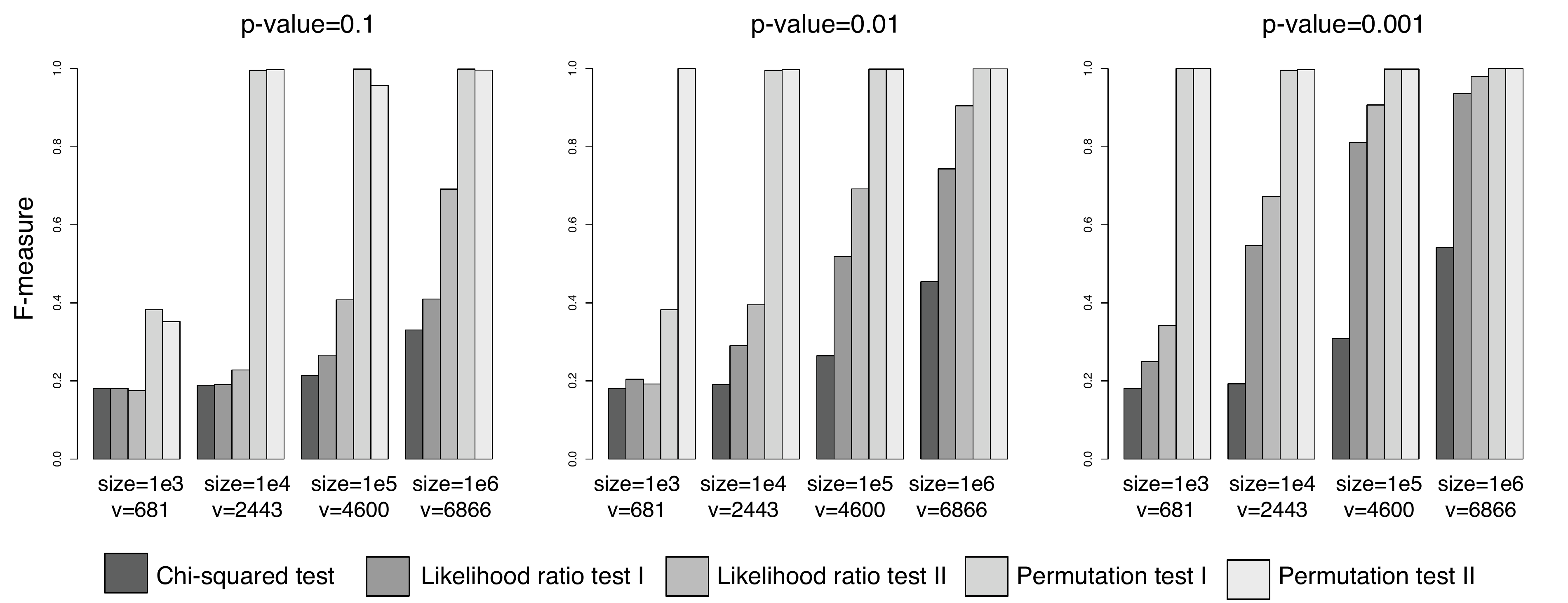}
   \caption{\label{fig:f-measure} The F-measure for simulated corpora
     of different sizes ($10^3$, $10^4$, $10^5$, and $10^6$ words) and
     for three different p-values.  Likelihood ratio test I is the
     method of \cite{Dunning:1993}.  Likelihood ratio test II is the
     back-off model in \mysec{model} using the asymptotic distribution
     of the log likelihood ratio.  Permutation test I is the
     multinomial model of~\cite{Pedersen:1996}.  Permutation test II
     is the procedure of \mysec{model}.  All permutation tests used
     $1000$ permutations.  Methods relying on the asymptotic
     distribution of the test statistic perform better as more data is
     seen.  Methods that employ the permutation test perform well on
     all data set sizes, and perform better than those methods relying
     on asymptotics.  For this simulated data, the model of
     \mysec{model} performs as well as a simple multinomial
     model. However, it further allows for finding multi-word
     expressions within a proper language model.  (See~\myfig{huff}).}
\end{figure*}

\label{sec:results}

We first compare the permutation test for back-off models with
previous methods of finding significant bigrams.  We then demonstrate
the full procedure for visualizing topics.

\subsection{Simulated bigrams}

We evaluate our method on simulated text data with known bigrams.  The
data are drawn sequentially from a Chinese restaurant process
(CRP)~\citep{Pitman:2002}.  The CRP is a distribution over a
potentially infinite vocabulary.  To simulate $N$ words, we draw each
from a distribution where the probability of any previously seen word
is proportional to the number of times it has been drawn, and the
probability of a new word is proportional to a scaling parameter
$\alpha$.  More formally, the $n$th word is drawn from the following
distribution,
\begin{equation*}
  p(w_{n+1} = v \g w_{1:n}) \propto
\begin{cases}
    n_v & \textrm{if $v$ exists;}\\
    \alpha & \textrm{if $v$ is a new word.}
\end{cases}
\end{equation*}
We embellish the CRP to create a corpus with bigrams.  When a new term
is to be added to the vocabulary, it will be a collocation of two
previously existing singleton terms with probability $\beta$.  It will
be a new singleton term with probability $1-\beta$.  CRP-based
distributions such as this one have been shown to match qualities of
word frequencies found in natural language~\cite{Goldwater:2006a}.

Simulating from this process yields a random corpus with a vocabulary
containing both singletons and bigrams.  However, an observed bigram
is indistinguishable from a pair of singletons.  Thus, using this corpus
as input to a word collocation algorithm, we can compare the set of
found bigrams to the true set of bigrams.  We measure success with
weighted precision and recall.  Note that in these simulations we did
not find or produce phrases of more than two words, as our purpose is
to compare to previous techniques.

We compared several tests for bigram discovery.

\textbf{The $\chi^2$ test.}  This is a classical test of
independence for discrete variables.

\textbf{Likelihood ratio tests.}  Here, we obtain p-values from
the asymptotic distribution of twice the likelihood ratio.  We
implemented a simple multinomial model~\citep{Dunning:1993} and the
back-off model in \mysec{model}.

\textbf{Permutation tests.}  As described above, these yield a
distribution-free method of obtaining p-values.  We employed a
permutation test with both a simple multinomial
model~\citep{Pedersen:1996} and the back-off model from \mysec{model}.
(A comparative study was not performed in~\cite{Pedersen:1996}.)

\myfig{f-measure} illustrates the F-measure achieved by these tests
for four simulated corpora of different sizes and for three p-value
thresholds.  The corpora were created with parameters $\alpha=1000$
and $\beta=0.1$.  The tests that rely on asymptotics improve
performance as the corpus size increases.  Permutation tests perform
well on small and large corpora.  The simple multinomial model is as
effective as the model of \mysec{model} for this task, but the
procedure presented here allows for recursive detection of word
phrases.



\subsection{Example topic visualizations}
\label{sec:topics}

\begin{figure*}[t]
\begin{center}
\includegraphics[width=\textwidth]{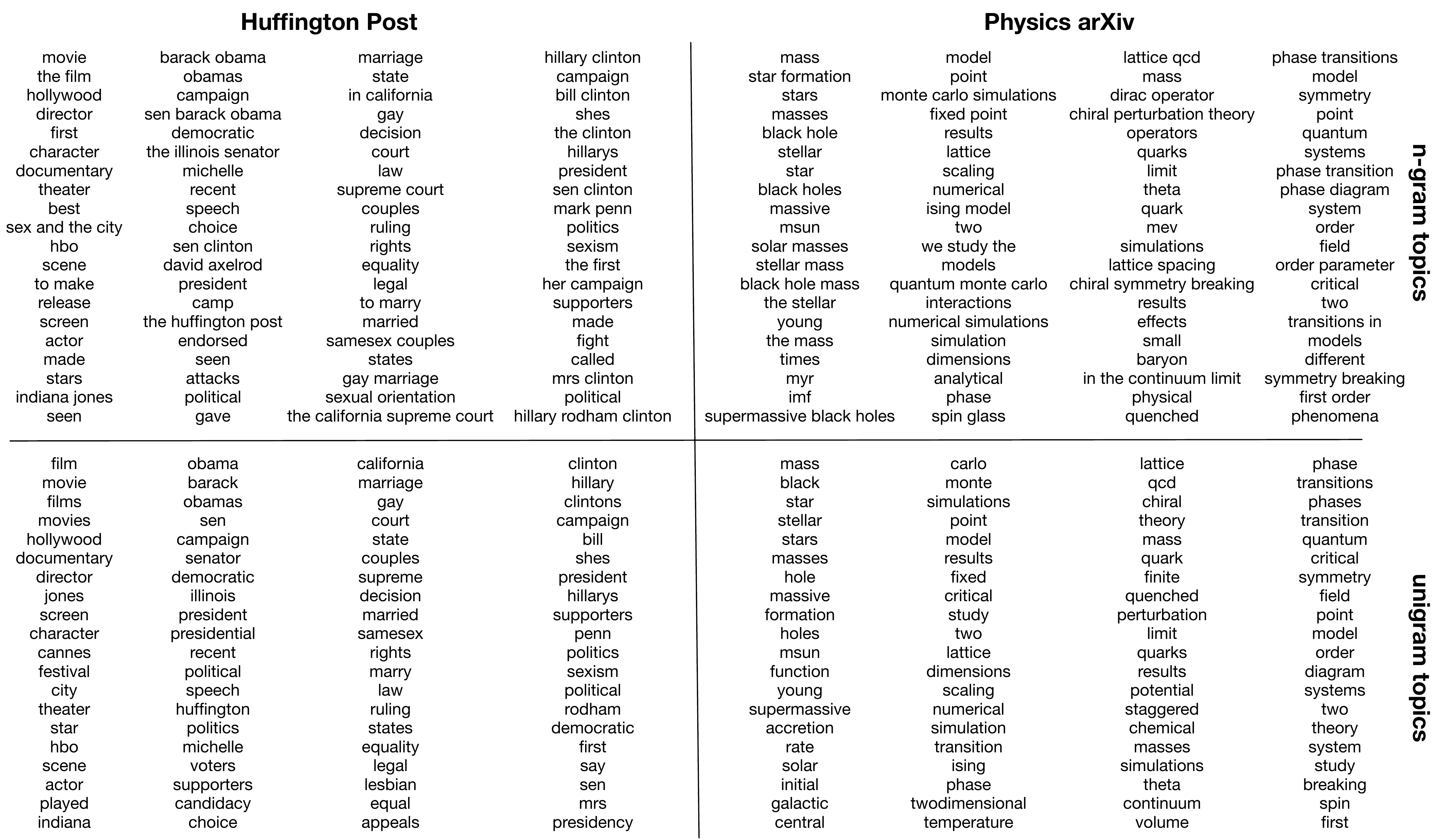}
\caption{\label{fig:huff} Standard unigram display of topics compared
  with turbo topics for two corpora, the Huffington Post (left), and
  physics arXiv (right).  Four topics are shown for each corpus,
  comparing the unigram visualization (bottom) with the turbo topic
  visualization (top).  The presentations that include the $n$-grams
  are more descriptive, uncovering $n$-grams such as ``indiana jones''
  and ``the california supreme court'' in the case of the Huffington
  Post, and ``monte carlo simulations'' and ``chiral symmetry
  breaking'' in the case of the physics abstracts.}
\end{center}
\end{figure*}


We demonstrate turbo topics with two corpora.\footnote{Code can be
  found at http://www.cs.princeton.edu/$\sim$blei/.}  First, we find
topics from each corpora using variational expectation
maximization~\cite{Blei:2003b}.  We then restrict our attention to
contexts surrounding the words assigned to each topic, and use the
recursive permutation tests of \mysec{strategy} to find a set of
back-off models, one for each topic.

To visualize, we consider the significant phrases of each model, i.e.,
those $n$-grams for which the algorithm chose to explicitly represent
parameters.  We order the $n$-grams by their probabilities.  To aid in
visualization, when an $n$-gram subsumes a shorter $n$-gram with lower
probability, we incorporate the shorter $n$-gram's mass into the
longer $n$-grams probability.  (The shorter $n$-gram's probability is
determined by how often it occurs without being part of the longer
$n$-gram.)  For example, if an expanded topic contains ``The New York
Mets'' with high probability and ``New York Mets'' with lower
probability, then we add the probability of the shorter phrase to the
longer phrase.  If a shorter phrase, such as ``court,'' appears on its
own with higher probability than a longer phrase, such as ``supreme
court,'' then both are considered.




Our first corpus contains articles from the Huffington Post, an online
news service.  This corpus contains 4000 documents and has a
vocabulary of 6500 terms.  Second, we use the 2006 physics abstracts
from arXiv.org, an online scientific preprint service.  This corpus
contains 50,000 documents and has a vocabulary of 17,000 terms.  For
both corpora, the vocabulary was obtained by removing stop words and
infrequent terms (appearing in fewer than 20 documents) from the topic
analysis.  These terms were, however, considered in the phrase
analysis.

Figure~\ref{fig:huff} shows, for each collection, four of the original
topics and the corresponding turbo topics.  (The news model contains
100 topics; the physics model contains 200 topics.  For both corpora,
the number of permutations was 100 and the p-value threshold was
0.01.)  Under the expanded view with bigrams and longer phrases, what
the topic is ``about'' comes into sharper focus.  For instance, while
we see from the ordered list of unigrams that a topic is about movies,
it may not be immediately apparent what ``jones'' and ``city'' refer
to.  In the expanded visualization, the phrases ``indiana jones'' and
``sex in the city'' provide a clearer indication why these terms
appear with such high probability.  Similarly, in a topic that
concerns gay marriage, ``the california supreme court'' appears,
offering a refinement of the terms ``court'' and ``supreme'' which are
separated in the standard probability-sorted unigram list.  Similar
effects are seen in the topics extracted from the corpus of physics
abstracts.  While one of the unigram topics assigns high weight to
``black'' and ``holes,'' the phrases ``black hole mass,'' ``star
formation,'' and ``supermassive black holes'' are more suggestive of
how the topic is used.


\section{Discussion}

Topic models are formulated and estimated based on the assumption of
word-level exchangeability, which leads to relatively simple and
computationally efficient inference algorithms.  This ``bag of words''
assumption is reasonable for identifying topics, but it becomes a
handicap when interpreting them.  The salient bigrams and phrases for
a topical word provide an indication of the role it plays in the
topic.

We have developed a new procedure for determining the salient phrases
for a topic.  Our procedure preserves the simplicity of an
exchangeable model while incorporating some of the context of richer
models.  Though we focused on topics derived from LDA, we emphasize
that this procedure can be used for any topic model, provided there is
a latent topic assignment variable for each word of the corpus.  Other
examples include author-topic models~\citep{Rosen-Zvi:2004} and
conditional topic models~\citep{Mimno:2008}.  Moreover, although we
focused on topic visualization, one can imagine other uses of the
resulting phrases, e.g., for information retrieval, that would not
require a full generative model.




In terms of statistical methodology, our results demonstrate that the
use of permutation testing is appropriate and effective in this
setting, where sparse statistics render tests based on asymptotic
distributions less accurate.  Although we have implemented a simple
greedy strategy based on recursively applying the permutation test, it
would be of interest to investigate more computationally efficient
procedures for simultaneously testing and correcting for multiple
hypotheses for expanding word phrases.

\section*{Acknowledgments}

We thank Jason Eisner, Chris Genovese, Noah Smith, and Larry Wasserman
for useful discussions about this work.

\bibliographystyle{plain}
\bibliography{bib}

%


\end{document}